\documentclass[preprint,12pt]{elsarticle}

\usepackage{amssymb}

\journal{Pattern Recognition}

\newcommand{\etal}{\emph{et al.}}
\usepackage{booktabs}
\usepackage{multirow}
\usepackage{graphicx}
\usepackage{amsmath}
\usepackage{amsfonts}
\usepackage{enumitem}

\usepackage{makeidx}
\usepackage{color}
\usepackage{tikz}
\usepackage{subfig}
\usepackage{graphics}

\usepackage[usestackEOL]{stackengine}
\usepackage[normalem]{ulem}
\newcommand{\minus}{\scalebox{0.75}[1.0]{$-$}}

\usepackage{textcomp}

\begin{document}
\begin{frontmatter}



\title{Candidate Fusion: Integrating Language Modelling into a Sequence-to-Sequence Handwritten Word Recognition Architecture}


\author{Lei Kang$^{* \dag}$, Pau Riba$^{*}$, Mauricio Villegas$^{\dag}$, Alicia Forn\'{e}s$^{*}$, Mar\c{c}al Rusi{\~n}ol$^{*}$  \\
$^{*}$Computer Vision Center, Barcelona, Spain\\
{\tt\small \{lkang, priba, afornes, marcal\}@cvc.uab.es}
\\
$^{\dag}$omni:us, Berlin, Germany\\
{\tt\small \{lei, mauricio\}@omnius.com}
}


\begin{abstract}
Sequence-to-sequence models have recently become very popular for tackling handwritten word recognition problems. However, how to effectively integrate an external language model into such recognizer is still a challenging problem. The main challenge faced when training a language model is to deal with the language model corpus which is usually different to the one used for training the handwritten word recognition system. Thus, the bias between both word corpora leads to incorrectness on the transcriptions, providing similar or even worse performances on the recognition task. In this work, we introduce Candidate Fusion, a novel way to integrate an external language model to a sequence-to-sequence architecture. Moreover, it provides suggestions from an external language knowledge, as a new input to the sequence-to-sequence recognizer. Hence, Candidate Fusion provides two improvements. On the one hand, the sequence-to-sequence recognizer has the flexibility not only to combine the information from itself and the language model, but also to choose the importance of the information provided by the language model. On the other hand, the external language model has the ability to adapt itself to the training corpus and even learn the most commonly errors produced from the recognizer. Finally, by conducting comprehensive experiments, the Candidate Fusion proves to outperform the state-of-the-art language models for handwritten word recognition tasks.

\end{abstract}


\end{frontmatter}



\section{Introduction}
\label{sec:intro}

Handwritten word recognition is the computer vision task that provide computers the ability to read handwritten text from images. Handwritten content is found in volume in both historic document collections~\cite{romero2013esposalles}, but also on current administrative documents~\cite{moysset20192d} such as invoices, tax forms, notes, accident claims, etc. Automatic reading systems are particularly interesting for document digitization processes where paper documents are converted into machine encoded text. The information contained in such converted documents can be thus leveraged and used in any computer application, such as automatic decision making processes, document classification, automatic routing, etc. Unlike the recognition of typewritten text, handwritten word recognition is still a challenging research problem because of the large variability across different handwriting styles~\cite{crettez1995set}. In the last few years, with the rise of deep learning architectures, some handwritten word recognition applications have started to reach a satisfying performance in some specific and restricted use cases~\cite{mor2018confidence, pham2014dropout, bluche2017scan}. However, we are still far away from having a generic and robust system able to read any handwritten text.

Automatic decoding textual information in images have several particularities. On the one hand, text is sequential in nature, coming in left to right order in latin languages. Most of the state of the art approaches are based on recurrent architectures~\cite{puigcerver2017multidimensional, bluche2014a2ia} for leveraging this sequential information. On the other hand, text follows a particular set of syntactic rules and presents a well defined morphological structure. Text recognition systems often integrate statistical language models~\cite{marti2001using, krishnan2018word,dutta2012robust} that are able to complement the optical part boosting the overall recognition performance. Language models for handwritten word recognition implemented as probability distributions over sequences of characters and words, aim to provide context to discern between sequences of characters that that might look similar, intending to resolve ambiguities from the optical recognition part. Different language model approaches have been proposed in the literature, from n-grams~\cite{bunke2004offline} to neural network architectures~\cite{zamora2014neural}.

However, in most of the state-of-the-art handwritten word recognition systems, including the recent sequence-to-sequence-based ones~\cite{sueiras2018offline,kang2018convolve}, the optical recognition part and the language models are seen as two separate and independent modules that are trained separately. Each of those modules are optimized separately using different data corpora, images of handwritten text on the one side, and a separate text corpora used to train the language statistics on the other. At the inference time, both modules are combined together. In that sense, language models are used either as a post-processing step, aiming at correcting recognition errors with the most likely sequence of characters~\cite{hori2017multi, toshniwal2018comparison}, or as an integrated guiding module, steering the decoding process towards the best fitting letter succession~\cite{gulcehre2015using}.

In this paper we present a novel sequence-to-sequence-based handwriting recognition architecture that integrates the language model within the recognizer. Since the language model and the optical recognition parts are jointly trained and optimized, the language model does not just encode statistics about the language, but also models the most commonly produced errors from the optical decoder and how to correct them. 

\begin{figure}
\centering
\begin{tabular}{ccc}
\toprule
\includegraphics[width=2.3cm,height=.75cm]{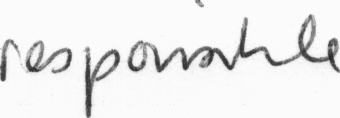}&
\includegraphics[width=2.3cm,height=.75cm]{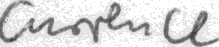}&
\includegraphics[width=2.3cm,height=.75cm]{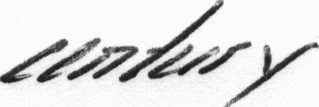}\\
\small{\texttt{hesporuaklly}} & \small{\texttt{ee}} &\small{\texttt{enterr}}\\
$\downarrow$& $\downarrow$& $\downarrow$\\
\small{\texttt{resporishle}} & \small{\texttt{Curluce}} &\small{\texttt{unterry}}\\
$\downarrow$& $\downarrow$& $\downarrow$\\
\small{\texttt{\textbf{responsible}}} & \small{\texttt{\textbf{currence}}}  & \small{\texttt{\textbf{century}}}\\

\midrule
\includegraphics[width=2.3cm,height=.75cm]{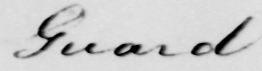}&
\includegraphics[width=2.3cm,height=.75cm]{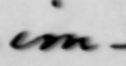}&
\includegraphics[width=2.3cm,height=.75cm]{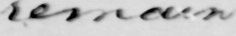}\\
\small{\texttt{were}} & \small{\texttt{ma}}  & \small{\texttt{eeree}}\\
$\downarrow$& $\downarrow$& $\downarrow$\\
\small{\texttt{suard}} & \small{\texttt{cim}}  & \small{\texttt{remann}}\\
$\downarrow$& $\downarrow$& $\downarrow$\\
\small{\texttt{\textbf{guard}}} & \small{\texttt{\textbf{im}}} & \small{\texttt{\textbf{remain}}}\\

\midrule
\includegraphics[width=2.3cm,height=.75cm]{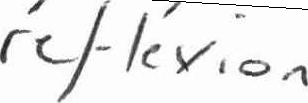}&
\includegraphics[width=2.3cm,height=.75cm]{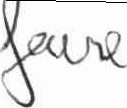}&
\includegraphics[width=2.3cm,height=.75cm]{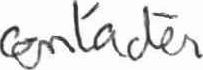}\\
\small{\texttt{RAlexiin}} & \small{\texttt{foure}}  & \small{\texttt{Cemad\'{e}r}}\\
$\downarrow$& $\downarrow$& $\downarrow$\\
\small{\texttt{r\'{e}fl\'{e}vion}} & \small{\texttt{feure}}  & \small{\texttt{contaderr}}\\
$\downarrow$& $\downarrow$& $\downarrow$\\
\small{\texttt{\textbf{r\'{e}flexion}}} & \small{\texttt{\textbf{faire}}} & \small{\texttt{\textbf{contacter}}}\\
\bottomrule
\end{tabular}
\caption{Transcription examples from the IAM, GW and Rimes datasets respectively. First row: training with only synthetically generated word samples; second row: fine-tuning with real training data; third word: joint training with the proposed language model (in boldface).}\label{fig:qualitative_samples}
\end{figure}

The handwriting word recognition architecture proposed in this paper significantly extends our preliminary work~\cite{kang2018convolve} by integrating a language model step within a sequence-to-sequence architecture. By incorporating the use of synthetic fonts and data augmentation strategies, we demonstrate the effectiveness and generality of our proposed approach in a significant amount of different public datasets and real industrial use cases. We exemplify in Figure~\ref{fig:qualitative_samples} the different transcription results that we are able to obtain with the proposed architecture.

The rest of the paper is organized as follows. In Section~\ref{sec:relate}, the state-of-the-art in handwritten word recognition is discussed. In Section~\ref{sec:data}, we present the data augmentation and pre-training processes leveraging the use of synthetic handwriting-looking data. In Section~\ref{sec:s2s}, our attention-based sequence-to-sequence model for handwritten word recognition is described. In Section~\ref{sec:lm}, we will focus on the proposed candidate fusion language model, while comparing it to two popular language models. In Section~\ref{sec:exp}, the datasets, the full experimental setup, the ablation study, and the results on popular handwriting datasets will be discussed in detail. Lastly, the conclusion is given in Section~\ref{sec:con}.

\section{Related Work}
\label{sec:relate}

Recognizing handwritten text has typically been addressed by combining computer vision and sequence learning techniques. The first handwritten word recognition approaches were based on Hidden Markov Models (HMMs)~\cite{bianne2011dynamic,bluche2013tandem,espana2010improving,gimenez2014handwriting}. Such approaches used to be successful pioneers, while nowadays, they have been outperformed by Neural Networks-based architectures. 
With the rise of neural networks, Recurrent Neural Networks (RNNs)~\cite{lipton2015critical} have started to become popular to deal with sequential data such as handwriting. For example, Bidirectional Long Short-Term Memory (BLSTM)~\cite{graves2008novel} or Multidimensional Long Short-Term Memory (MDLSTM)~\cite{graves2009offline} have been widely adopted by handwritten word recognition community. Lately, these models have been discussed and improved. For example, Puigcerver~\cite{puigcerver2017multidimensional} compared 1D-LSTM and 2D-LSTM layers to prove that multidimensional recurrent layers may not be necessary to achieve good accuracy for handwritten word recognition.  Toledo~\etal~\cite{toledo2017handwriting} provided an approach that combined character embeddings  with a BLSTM decoding. Most of the handwritten word recognition approaches today are based on the use of a recurrent network with Connectionist temporal classification (CTC) layers~\cite{graves2006connectionist}. However, for obtaining a reasonable performance on handwritten word recognition tasks, the CTC-based approaches have to stay along with its limitations, such as the high complexity and the slowness to train. The CTC module has become the bottleneck in sequence-based approaches. 

Recently, inspired by machine translation~\cite{bahdanau2014neural,sutskever2014sequence}, speech recognition~\cite{bahdanau2016end,chorowski2015attention} and image captioning~\cite{xu2015show}, the sequence-to-sequence architecture~\cite{sueiras2018offline,kang2018convolve} has started to be applied into handwritten word recognition tasks. These sequence-to-sequence approaches follow the architecture of encoder, decoder and attention mechanism. They present the advantage that by decoupling encoder and decoder, the output size is not determined by the input image width, so that the use of CTC can be avoided. For example, Sueiras~\etal~\cite{sueiras2018offline} provided a sequence-to-sequence based handwriting recognizer, but they imposed a manually set sliding-window. Our previous work \cite{kang2018convolve} analyzed various strategies to find a proper sequence-to-sequence based architecture for specifically targeting handwritten word recognition tasks. 

However, these sequence-to-sequence based handwriting recognizers do not integrate a language model in the whole system. Since the age of HMMs, there have been plenty of researches on the usage of linguistic knowledge to assist a HMM-based handwriting recognition process~\cite{marti2001using,perraud2003n,vinciarelli2003offline}. Later on, as the RNN-CTC model became the state-of-the-art on handwritten word recognition tasks, how to effectively integrate a language model  into a recognizer has been a hot topic concurrent with the development of a handwriting recognizer. For instance~\cite{bluche2014a2ia,moysset2014a2ia,bluche2014comparison,liu2015study,voigtlaender2016handwriting,puigcerver2017multidimensional} have integrated character n-grams language modelling into a RNN-CTC based handwriting recognizer. However, the n-gram model is just statistics on the co-ocurrence of characters computed over a text corpus and, even if they are helpful as an error-correcting post-processing step, they do not represent inherent language knowledge.

Concurrently, Recurrent Neural Network Language Models (RNNLM)~\cite{mikolov2010recurrent,gulcehre2015using,hori2017multi,toshniwal2018comparison} have been developed prosperously among machine translation and speech recognition communities, because they can learn an effective representation of variable length characters and memory a long enough character history,  outperforming the n-grams. Especially, Gulcehre~\etal~\cite{gulcehre2015using} provided two approaches: Shallow Fusion and Deep Fusion, which have been widely used and are the state-of-the-art RNNLMs in machine translation and speech recognition tasks. However, these RNNLMs are integrated into the sequence-to-sequence recognizer in a serial way. Both the sequence-to-sequence recognizer and the language models are trained separately and combined together in the final step. In that sense, the two different modules cannot properly benefit one from another and leverage the mutual information that both the optical recognizer and the language models could provide one to another. Our proposed candidate fusion language modelling is based on the idea that the optical part and the statistical character modelling shall communicate between each other, being able to jointly decode the most likely and most visually suitable character sequence.

\section{Getting Enough Training Data}\label{sec:data}

The first extension that we propose over our previous sequence-to-sequence architecture~\cite{kang2018convolve} is related to the training data pre-processing steps, namely the data augmentation step and the use of synthetic fonts.

\begin{figure*}
\centering
\begin{tabular}{c}
    \includegraphics[height=0.7cm]{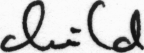}
    \includegraphics[height=0.7cm]{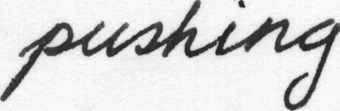}
    \includegraphics[height=0.7cm]{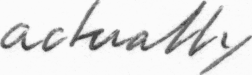}
    \includegraphics[height=0.7cm]{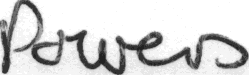}
    \includegraphics[height=0.7cm]{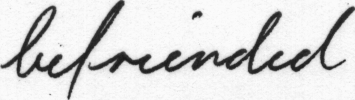}
    \includegraphics[height=0.7cm]{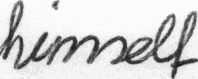}
    \\
    \includegraphics[height=0.7cm]{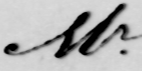}
    \includegraphics[height=0.7cm]{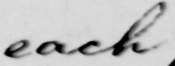}
    \includegraphics[height=0.7cm]{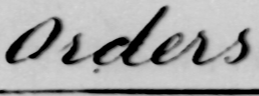}
    \includegraphics[height=0.7cm]{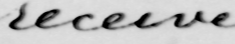}
    \includegraphics[height=0.7cm]{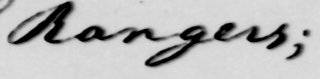}
    \includegraphics[height=0.7cm]{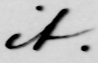}
    \\
    \includegraphics[height=0.7cm]{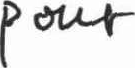}
    \includegraphics[height=0.7cm]{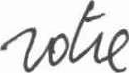}
    \includegraphics[height=0.7cm]{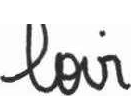}
    \includegraphics[height=0.7cm]{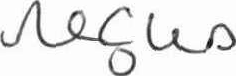}
    \includegraphics[height=0.7cm]{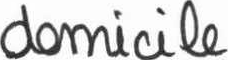}
    \includegraphics[height=0.7cm]{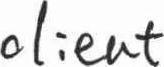}
    \includegraphics[height=0.7cm]{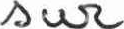}
\end{tabular}

\caption{Real word samples in IAM, GW and Rimes datasets, from top to bottom, respectively. Each example has different characteristics such as shear, stroke width, language, etc.}
\label{fig:real_words}
\end{figure*}

A system able to effectively recognize handwritten words should be able to deal with the inherent deformations of handwriting text. These deformations not only come from the different writing styles across different individuals, but also in words written by the same person at different times. Figure~\ref{fig:real_words} presents several real word images coming from different datasets and authors showing the huge variability in styles. Traditionally, to allow handwritten word recognition systems to generalize and prevent over-fitting, without having to manually annotate millions of word samples, data augmentation has been proposed in the literature. However, this technique is not able to increase the number of handwriting styles in the dataset. To solve this lack of diversity of handwriting styles, pre-training the recognition models with synthetic data is proposed. Intuitively, feeding more data that looks ``realistic'' as a pre-training provides a pre-condition to our system, making it able to extract the general features required for handwriting recognition. Afterwards, a fine tuning with real data will adapt it to the desired use case. In this Section, both techniques are presented and adapted to handwritten words.

\subsection{Data Augmentation}
Having enough training data is crucial for the performance of deep learning frameworks. To tackle this problem, some data augmentation techniques have been proposed in the literature. Usually, random image transformations are applied to the training data in order to increase the diversity. In our sequence-to-sequence setting, these transformations are constrained to obtain a realistically looking image where the text is readable. In this work, we specially designed a pipeline able to capture the variability of real data in the document domain. These set of operations with random parameters are applied among all epochs and consist of a blurring / sharpening step, elastic transformations by using a mesh grid~\cite{simard2003best}, shear, rotation, translation and scaling transforms, gamma correction and blending with synthetically generated background textures.

\begin{figure*}
\begin{tabular}{c}
     \includegraphics[height=6cm]{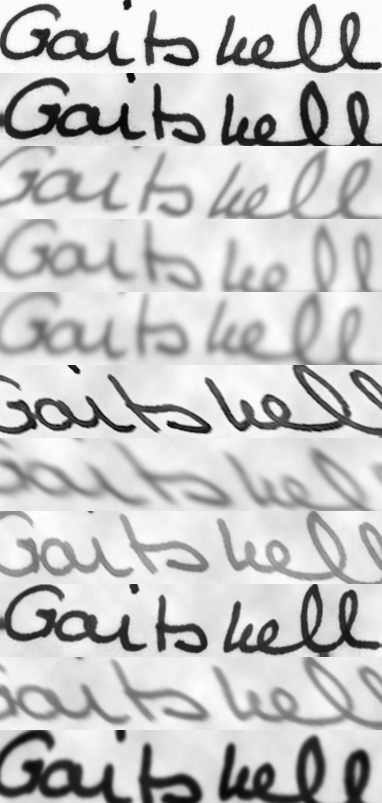}
     \includegraphics[height=6cm]{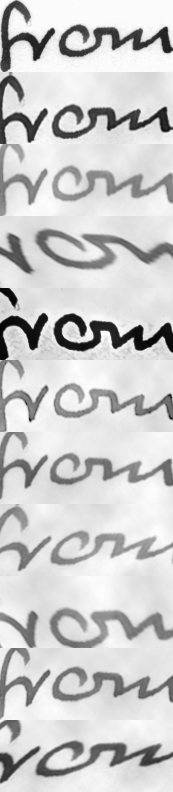}
     \includegraphics[height=6cm]{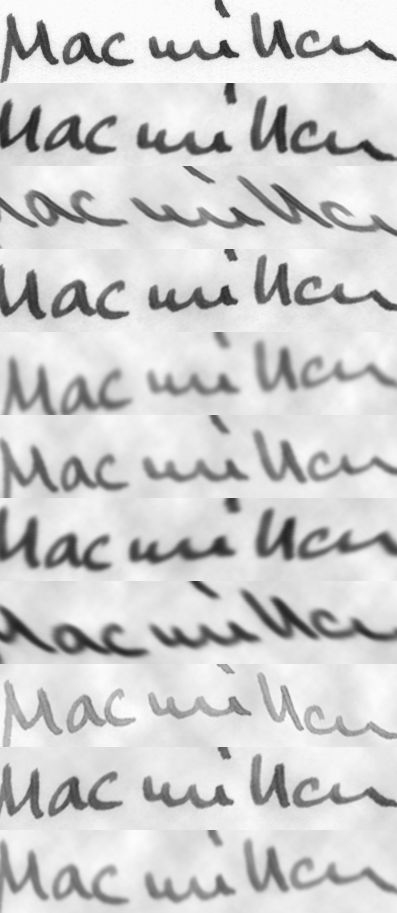}
     \includegraphics[height=6cm]{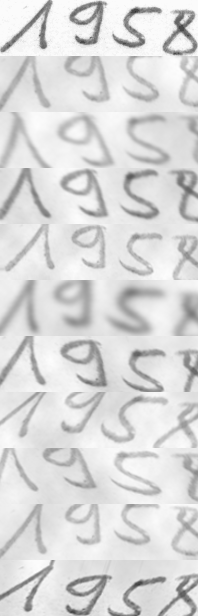}
     \includegraphics[height=6cm]{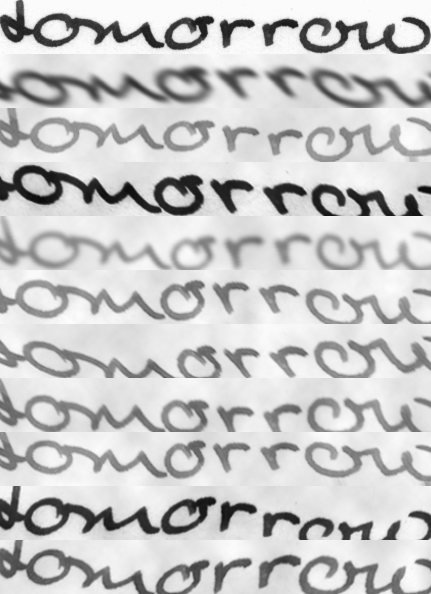}
\end{tabular}
\caption{Examples of data augmentation on real handwritten words. The first row shows real word samples, then followed by 10 variations of such word after random data augmentation.}
\label{fig:aug}
\end{figure*}

Figure~\ref{fig:aug} shows some examples that are used in training after the data augmentation module. Notice that the proposed operations introduce variations of word samples during training. This diversity helps  to some extend to prevent over-fitting and leads to models that are able to generalize better than training just with the original set of images. However, the generated words are restricted to a fixed lexicon and the writing styles provided by the training set. Hence, the system is not able to extend the vocabulary which is a key feature in handwritten word recognition systems.

\subsection{Pre-training with Synthetic Data}
\label{sec:syn}
Recently, it has become a common trend the use of synthetically generated images to magnify the training data volume~\cite{Dosovitskiy17}. Instead of generating realistic images, the idea is to encode the necessary information required for a desired task. Available public datasets, such as the IIIT-HWS dataset~\cite{krishnan2016generating}, have already tackled the generation of synthetically generated handwriting word collections by the use of truetype electronic fonts. Such approach has the advantage that one can virtually generate an infinity of annotated training samples for free. However, the available datasets do not consider special characters (e.g. accents, umlauts, punctuation symbols, etc.) that may be required. Hence, we defined our own data generator able to be used to train several languages taking into account its own peculiarities.

\begin{figure*}
\begin{tabular}{cccccc}
    Adversarial & after & included & Nester & embrouiller & tonnerre\\
    \includegraphics[height=0.55cm]{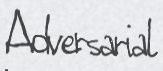}  &
    \includegraphics[height=0.55cm]{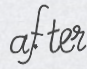}  &
    \includegraphics[height=0.55cm]{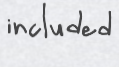}  &
    \includegraphics[height=0.55cm]{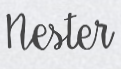}  &
    \includegraphics[height=0.55cm]{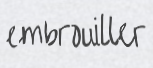}  &
    \includegraphics[height=0.55cm]{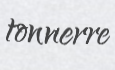}  \\
    
    \includegraphics[height=0.55cm]{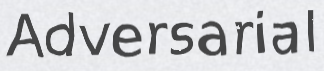}  &
    \includegraphics[height=0.55cm]{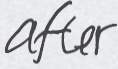}  &
    \includegraphics[height=0.55cm]{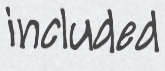}  &
    \includegraphics[height=0.55cm]{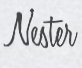}  &
    \includegraphics[height=0.55cm]{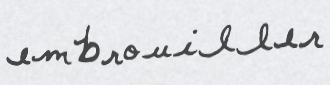}  &
    \includegraphics[height=0.55cm]{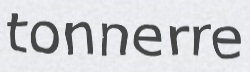}  \\
    
    \includegraphics[height=0.55cm]{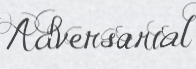}  &
    \includegraphics[height=0.55cm]{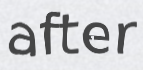}  &
    \includegraphics[height=0.55cm]{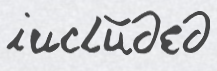}  &
    \includegraphics[height=0.55cm]{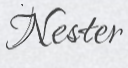}  &
    \includegraphics[height=0.55cm]{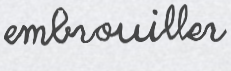}  &
    \includegraphics[height=0.55cm]{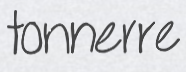}  \\
    
    \includegraphics[height=0.55cm]{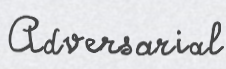}  &
    \includegraphics[height=0.55cm]{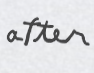}  &
    \includegraphics[height=0.55cm]{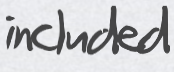}  &
    \includegraphics[height=0.55cm]{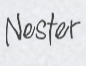}  &
    \includegraphics[height=0.55cm]{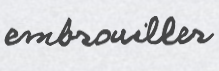}  &
    \includegraphics[height=0.55cm]{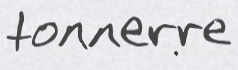}  \\
    
    \includegraphics[height=0.55cm]{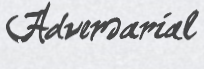}  &
    \includegraphics[height=0.55cm]{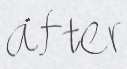}  &
    \includegraphics[height=0.55cm]{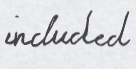}  &
    \includegraphics[height=0.55cm]{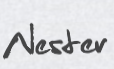}  &
    \includegraphics[height=0.55cm]{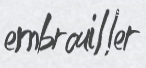}  &
    \includegraphics[height=0.55cm]{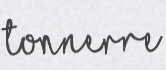}  \\
    
    \includegraphics[height=0.55cm]{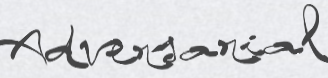}  &
    \includegraphics[height=0.55cm]{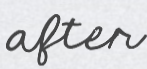}  &
    \includegraphics[height=0.55cm]{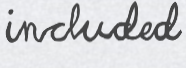}  &
    \includegraphics[height=0.55cm]{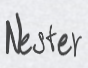}  &
    \includegraphics[height=0.55cm]{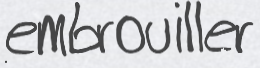}  &
    \includegraphics[height=0.55cm]{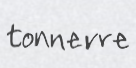}  \\
    
    \includegraphics[height=0.55cm]{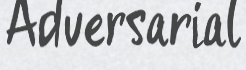}  &
    \includegraphics[height=0.55cm]{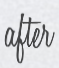}  &
    \includegraphics[height=0.55cm]{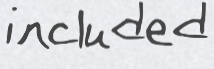}  &
    \includegraphics[height=0.55cm]{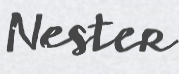}  &
    \includegraphics[height=0.55cm]{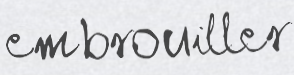}  &
    \includegraphics[height=0.55cm]{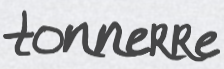}  \\
    
    \includegraphics[height=0.55cm]{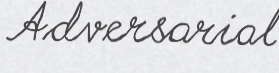}  &
    \includegraphics[height=0.55cm]{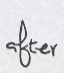}  &
    \includegraphics[height=0.55cm]{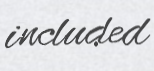}  &
    \includegraphics[height=0.55cm]{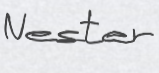}  &
    \includegraphics[height=0.55cm]{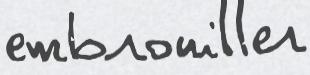}  &
    \includegraphics[height=0.55cm]{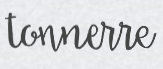}  \\
    
    \includegraphics[height=0.55cm]{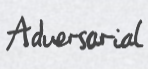}  &
    \includegraphics[height=0.55cm]{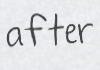}  &
    \includegraphics[height=0.55cm]{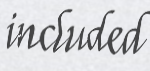}  &
    \includegraphics[height=0.55cm]{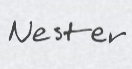}  &
    \includegraphics[height=0.55cm]{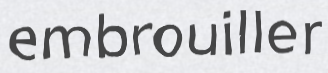}  &
    \includegraphics[height=0.55cm]{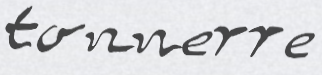}  \\
    
    \includegraphics[height=0.55cm]{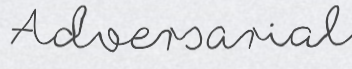}  &
    \includegraphics[height=0.55cm]{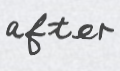}  &
    \includegraphics[height=0.55cm]{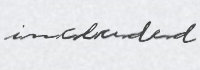}  &
    \includegraphics[height=0.55cm]{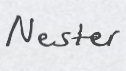}  &
    \includegraphics[height=0.55cm]{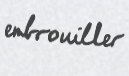}  &
    \includegraphics[height=0.55cm]{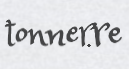}  \\
\end{tabular}
\caption{Examples of synthetic data generation. The first row is the given word from a public dictionary, then followed by 10 rendered image samples with different electronic fonts and random augmentation.}
\label{fig:syn_words}
\end{figure*}

As a text corpus, several books written in English and French have been used. These books are freely available on the Internet and will model the language character distribution. From these books, over 250.000 unique words were collected. Afterwards, we randomly render those words with 387 freely available electronic fonts that imitate cursive handwriting. However, for a given font, all of the instances of a character will always look the same. In order to overcome such limitation, the same data augmentation pipeline previously presented has been applied. This augmentation step is applied online within the data loader, so that each batch is randomly augmented. Some samples of synthetic words are shown in Figure~\ref{fig:syn_words}.

\section{Sequence-to-Sequence Word Recognizer}
\label{sec:s2s}
Our baseline handwritten word recognizer, previously introduced in~\cite{kang2018convolve} follows an encoder-decoder architecture. An attention mechanism is used to help the system focus at some spatial locations of the image when decoding character by character.

\begin{figure}[]
    \centering
    \includegraphics[width=0.9\linewidth]{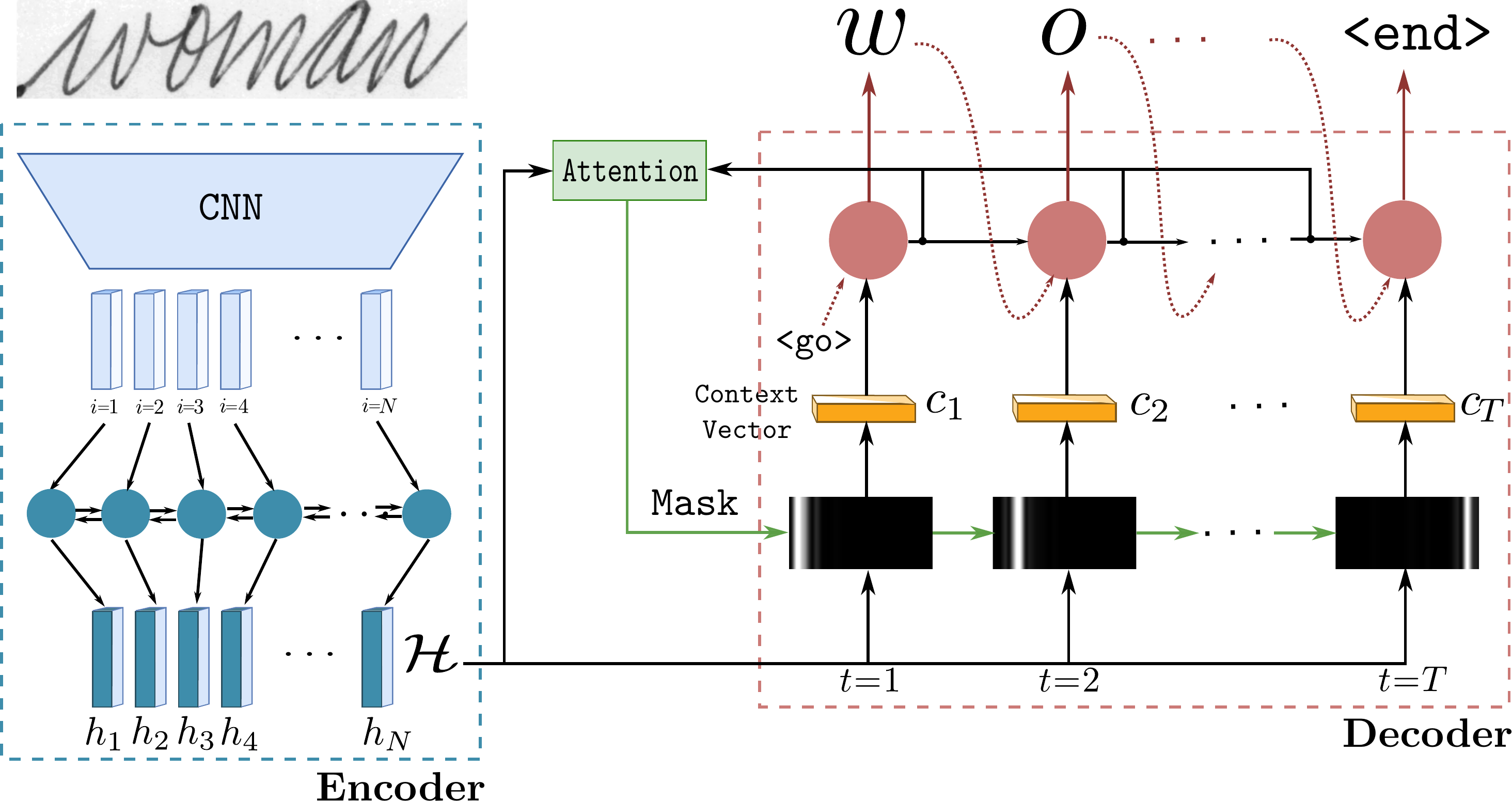}
    \caption{Architecture of the proposed sequence-to-sequence based recognizer.}
    \label{fig:arch}
\end{figure}

Figure~\ref{fig:arch} introduces the whole pipeline of the proposed sequence-to-sequence architecture. The proposed model has three components: encoder, attention and decoder. Each part will be detailed in the following sections.

\subsection{Encoder}
\label{sec:enc}

The first component of our architecture is an encoder, shown in the blue box in Figure~\ref{fig:arch}, whose objective is to extract high-level features from handwriting word images, that hopefully will discriminate between different glyphs. These features should be able to codify the contents of the image as well as the order. In this work, we tested two different approaches for the encoding part. On the one hand a Convolutional Neural Network with positional encoding~\cite{vaswani2017attention}, and, on the other hand a CNN followed by a Recurrent Neural Network architecture, specifically, a Bidirectional Gated Recurrent Unit (BGRU).

Both the positional encoding and the recurrent network modules play the role to provide positional information. The idea is to help the decoder to follow the proper order during the decoding process. Different CNN architectures will be further discussed in Section~\ref{sec:ablation}. We denote $h_i \in \mathcal{H}, i \in \{1,2,\dots,N\}$ as the output sequence of the encoder, where $\mathcal{H}$ is the final feature representation of the whole word and $N$ is the length of $\mathcal{H}$. Note that different word lengths will yield different feature sizes $N$. Figure~\ref{fig:arch} presents the encoder with a RNN.

\subsection{Attention Mechanism}
\label{sec:attn}
The basic idea of attention mechanisms is to focus the decoder part of the network to specific regions containing relevant information at each different time step. In our particular scenario, what we expect is that the attention mask focuses at each different character at each decoding step following the reading order. By using attention mechanisms, the decoder tasks becomes much more simple since it should only focus on individual character recognition rather than obtaining also the proper order and feature alignment.

In our work we compare and discuss two popular attention mechanisms applied to the specific task of recognizing handwritten words. A detailed performance evaluation is provided in Section~\ref{sec:ablation}.

\subsubsection{Content-based Attention}
Attention mechanisms were first proposed in the machine translation field to help the decoder in deciding which words from the source sentence we shall focus on to predict the proper output word in the target language. In this setting Bahdanau~\etal~\cite{bahdanau2014neural} proposed an attention model which deals with the content of the input features. In this case, the order is not taken into account in this model and should be processed in the encoder step either by using positional encoding or a recurrent architecture. This occurs because the proposed mechanism deals with the feature contents rather than using a combination with the previous attention mask.

In this setting, we define $\alpha_{t}$ as the mask vector of the attention at time step $t$ and $s_{t}$ as the hidden state of the
decoder at current time step $t\in\{1,2,\dots,T\}$, where $T$ is the maximum length of the transcription. The mask vector of the attention $\alpha_{t}$ is calculated by:

\begin{equation}
    \alpha_{t} = \operatorname{Softmax}(e_{t}) \label{equ:softmax}
\end{equation}
where $e_{t, i} = f(h_{i}, s_{t-1})$ is a learnable function. In this model $f(h_{i}, s_{t-1}) = w^T \tanh(Wh_{i} + Vs_{t-1} + b)$ where $w$, $W$, $V$ and $b$ are trainable parameters.


We denote as context vector the features that will be fed to the decoder at each time step. These features are generated from the attention mask $\alpha_t$ and the encoder feature representation $H$. Thus, the context vector for the corresponding character pixels is be obtained by,
\begin{equation}
	    c_{t} = g(\alpha_{t}, H) = \sum_{i=0}^{N\text{-}1} \alpha_{ti}h_{i}
\end{equation}

\subsubsection{Location-based Attention}
Despite the use of positional encoding or recurrent networks, it is still a challenging task to track the attention steps across the word image. In this sense, Chorowski~\etal~\cite{chorowski2015attention} introduced a new attention method that makes use of the previous attention mask. In practice, this is done by adding an extra location term $l_{t}$, which is calculated by:

\begin{equation}
	    l_t = F * \alpha_{t-1}
\end{equation}
where $F\in\mathbb{R}^{k\times r}$ is a matrix, $*$ is a convolutional operation. Thus, the location-based attention can be written as:
\begin{equation}
	    e_{t, i} = f'(h_{i}, s_{t-1}, l_{t}) = w^T \tanh(Wh_{i} + Vs_{t-1} + Ul_{t,i} + b)
\end{equation}
where $w$, $W$, $V$, $U$ and $b$ are trainable parameters. In this way, we have explicitly included the location information into the attention mechanism.

\subsection{Decoder}
\label{sec:dec}

Given the corresponding context vector and a $\langle go \rangle$ symbol, the decoder should be able to start the decoding process of the image text. The decoder is implemented as a unidirectional multi-layered GRU, which has enough capacity to predict a character at each time step. This character is then fed to the next iteration until an $\langle end \rangle$ symbol appears.

The decoder's output are the different predicted characters $\textbf{y}=\{y_{1},y_{2},\dots,y_{T}\}$ at each time step, where $T$ is the maximum length of the final transcription. Our proposed decoder unit is different from the conventional decoder unit that has been utilized in other sequence-to-sequence approaches as shown in the red box in Figure~\ref{fig:arch}. We have simplified the procedure between the decoder hidden state $s_{t}$ and the output logit $y_{t}$ at current time step $t$, because the decoder unit itself has enough ability to produce a proper character output. The comparison between both two architectures will be detailed in Section~\ref{sec:ablation}.

The predicted character at the current time step $t$ is calculated by:
\begin{equation}
	    y_{t} = \arg\max(\omega(s_{t})) 
\end{equation}
where $\omega(\cdot)$ is a linear operation to match the size of logit output to the size of possible characters. And then, to keep it simple and efficient, we just pick up the character with the highest probability and transfer it to a corresponding embedding vector $\tilde{y}_{t}$ by a embedding layer:
\begin{equation}
	    \tilde{y}_{t} = \operatorname{Embedding}(y_t)
	    \label{equ:embed}
\end{equation}

The input of each decoder unit consists of the previous prediction $\tilde{y}_{t-1}$ and the context vector $c_{t}$, so each hidden state of the decoder $s_{t}$ is calculated by:
\begin{equation}
	    s_{t} = \operatorname{Decoder}([ c_{t}, \tilde{y}_{t-1}], s_{t-1})
	    \label{equ:dec_state}
\end{equation}
where $[\cdot,\cdot]$ is the concatenation of two vectors.
   
At the beginning of the decoding process, we always feed into the start signal $\langle go \rangle$ as the first input character, while the prediction process ends when the end signal $\langle end \rangle$ occurs or until the maximum time step $T$ is reached.

\section{Candidate Fusion Language Model}
\label{sec:lm}
In this section, we propose a novel way to integrate language models into sequence-to-sequence models for handwritten word recognition tasks, that we coined as candidate fusion. The main idea is that the we first train a very simple language model with just text corpora (no images) with a recurrent neural network that given a sequence of characters is able to predict which is the most likely character to come next. This would be a similar idea of the well known word2vec models (e.g. skipgram) that are able to deduce most likely words within context, but for characters. Once this language model is pre-trained, now we can combine it with the optical decoder, so that the input to the decoder are not only the attended visual features at each particular time step, but also which is the most likely characters to be decoded given the ones that have been decoded so far.

\begin{figure*}
    \centering
    \setlength{\tabcolsep}{2em}
    \begin{tabular}{ccc}
        \includegraphics[height=4.5cm]{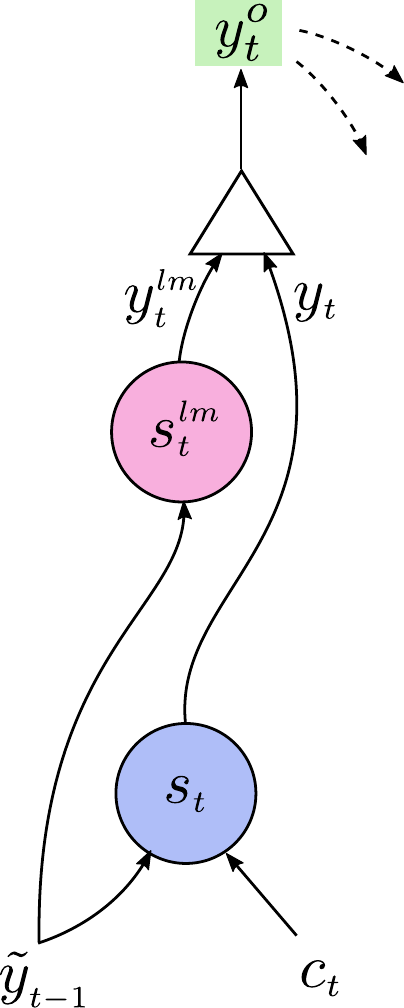} & \includegraphics[height=4.5cm]{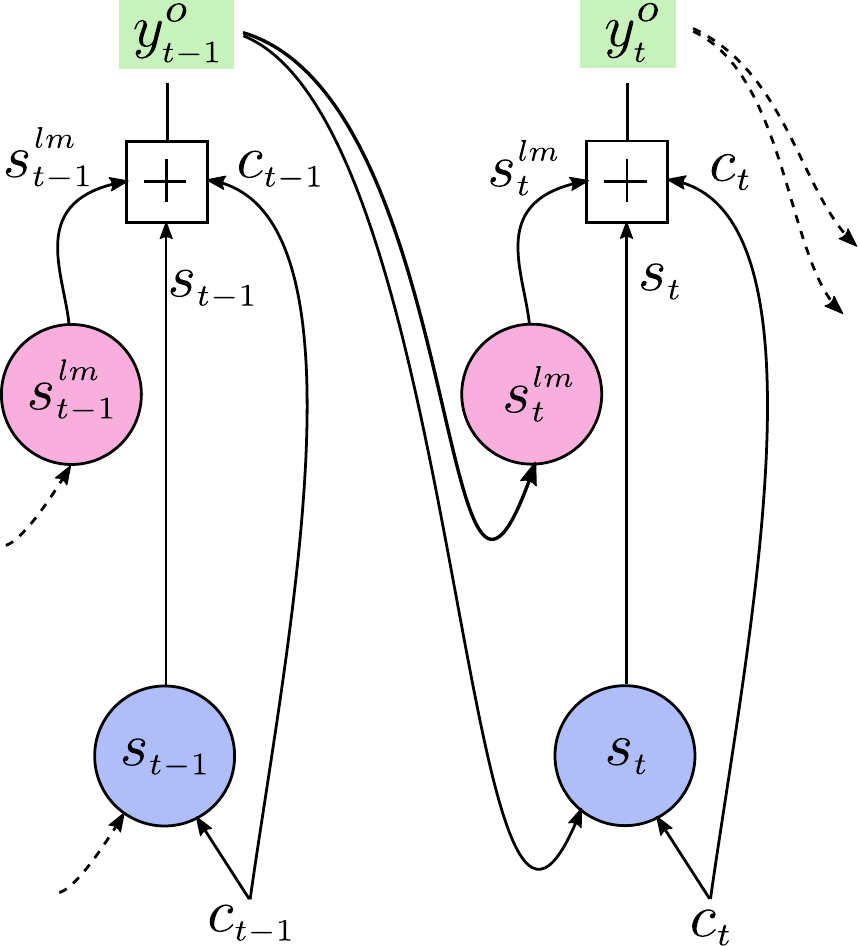} & \includegraphics[height=4.5cm]{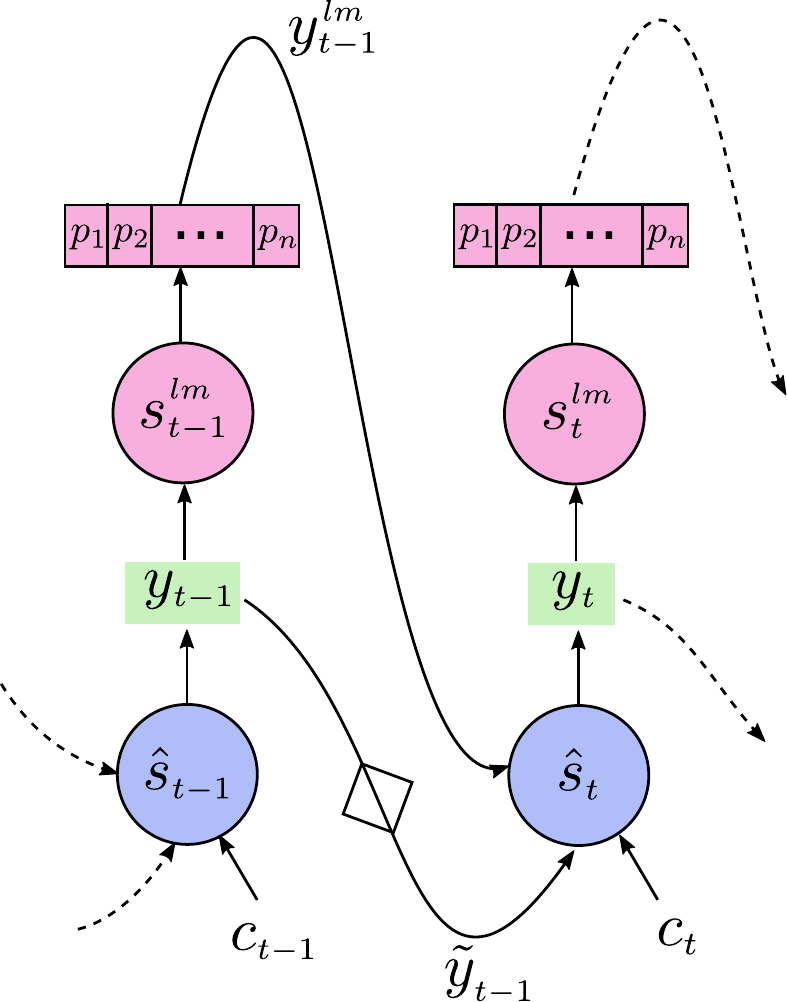}\\
        (a) & (b) & (c)
    \end{tabular}{}
    \caption{Architecture of the language models: (a) Shallow Fusion, (b) Deep Fusion, and (c) our proposed Candidate Fusion.}
    \label{fig:lm}
\end{figure*}

Unlike the popular Shallow Fusion and Deep Fusion language models~\cite{gulcehre2015using}, shown in Figure~\ref{fig:lm}(a) and (b) respectively, the final prediction is not decided by merging the outputs of the recognizer and the language model. The role of our language model is to provide other probabilities among all the characters, as indicated by $y_{t}^{lm}$, where $t$ is the current time step during the decoding stage. This language model information will be injected into the decoder as one of the inputs. So the new hidden state of the decoder $\hat{s}_{t}$ is calculated by:
\begin{equation}
	    \hat{s}_{t} = \operatorname{Decoder}([c_{t}, \tilde{y}_{t-1}, {p}_{t-1}^{lm}], \hat{s}_{t-1})
	    \label{equ:dec_state_lm}
\end{equation}
where ${p}_{t-1}^{lm}$ is the output of the language model from $s_{t-1}^{lm}$ at the previous time step $t\minus1$. The effect of the linguistic knowledge will be extensively analyzed in~Section~\ref{sec:ablation}.

The difference between Equations~\ref{equ:dec_state} and~\ref{equ:dec_state_lm} is that we add now a second ``adviser'' ${p}_{t-1}^{lm}$ into our decoder. Thus, the decoder can learn a trade-off between its output $\tilde{y}_{t-1}$ and that of an additional language model. We call it ``adviser'' because the decoder unit could choose to take into account the information from the ``adviser'' or totally ignore it. In this way, the explicit language model will never make the recognition performance worse than the baseline that is trained without language model at all. The reason is that, in an extreme case, if the bias of the language knowledge between the training data and the outside corpus is too high, the decoder can be adapted to predict transcriptions by ignoring the language model and just relying on the optical part. 

Shallow Fusion directly applies a language model to the final prediction of the decoder by simply summing up both the probabilities of the recognizer $y_t$ and the language model $y_t^{lm}$, as shown in Figure~\ref{fig:lm}~(a). Because of the bias of the language knowledge between the training corpus and the outside corpus, summing up the probabilities of both the recognizer and language model modules may produce incorrect final transcriptions. Therefore, to make full use of Shallow Fusion, one must carefully select a corpus that shares most of the words within the target dataset and tune the weight hyper-parameter that is in charge of the trade-off between both the probabilities of the recognizer and the language model.

Deep Fusion shares the same language model as Shallow Fusion, but it goes one step further to merge both information from the recognizer and the language model in the feature level as shown in Figure~\ref{fig:lm}~(b). The decoder of the recognizer and the language model are two independent pipelines, while the hidden state of the recognizer $s_{t}$, the hidden state of the language model $s_{t}^{lm}$ and the context vector $c_{t}$ at time step $t$ merge together by concatenating them. Afterwards, the merged feature goes through a fully connected layer and an activation layer to generate final prediction $y_t^o$. Both the recognizer and the language model contribute to the final prediction, but they are still independent from each other. Thus, this method still can not handle the bias of the language knowledge between both the training corpus and the outside corpus. In any case, as it can be jointly fine-tuned, the performance could be better than the Shallow Fusion case. 

Our Candidate Fusion language model, shown in Figure~\ref{fig:lm}~(c), is designed to further boost the performance. The intuition behind is to take advantage of the mutual information from both the optical recognizer and the morphology of the tackled language. This means that the decoder incorporates information both from the attended visual features and the language knowledge at each time step, and, at the same time, the language model itself can also adapt to the most common mistakes made by the recognizer. To do this, at each time step $t\minus1$, the language model takes the final prediction $y_{t-1}$ as input and outputs a corrected version $y_{t-1}^lm$ utilizing the learnt knowledge from the outside corpus. Then, at the next time step $t$, the recognizer takes the previous prediction of the recognizer $y_{t-1}$, the corrected version of the language model $y_{t-1}^lm$ and the current context vector $c_t$ as inputs to generate the final prediction $y_t$. At Figure~\ref{fig:lm} we can see our difference that the final prediction is taken from the recognizer and the language model is highly integrated into the recognizer system as a candidate prediction, that is why we denote this method Candidate Fusion. We believe that it is a natural way to integrate a language model. In Section~\ref{sec:exp} we will show the performance improvement on popular datasets.

\section{Experiments}
\label{sec:exp}
In this section we present the extensive evaluation of our proposed approach. First, we perform several ablation studies on the key components to analyze the most suitable architecture. Second, we compare our recognizer with the state-of-the-art models on handwritten word recognition. Next, we analyze the performance of the most popular language models and compare with the proposed candidate fusion approach. Further, we apply a simple edit-distance-based lexicon to evaluate how the use of a closed lexicon can boost the performance. Finally, we provide an experiment on an industrial use case.

\subsection{Datasets and Metrics}

We will use several datasets for the experimental evaluation. They have been selected because of their different particularities: single or multiple writers, modern or historical documents or written in different languages. The IAM, George Washington (GW) and Rimes datasets are publicly available, whereas CarCrash is a private dataset. The details of these datasets are shown in Table~\ref{tab:datasets}. 

The standard Character Error Rate (CER) and Word Error Rate (WER) metrics are utilized to evaluate the system's performance. Formally,

\begin{equation}
	   CER = \frac{S + I + D}{N}
\end{equation}
where $S$, $I$, $D$ are the number of character substitutions, insertions and deletions, respectively. $N$ is the total number of characters in the groundtruth transcription. 

\begin{equation}
	   WER = \frac{S_w + I_w + D_w}{N_w}
\end{equation}
The WER metric is computed similar to CER. In this case, $S_w$, $I_w$, $D_w$ and $N_w$ refer to words instead of characters. In the following experiments, the CER and WER metrics will range from [0-100]. Thus, a lower value indicates a better performance.

\begin{table}
    \centering
        \caption{Overview of the different datasets used in this work depicting its characteristics.}
    \label{tab:datasets}
    \footnotesize
    \begin{tabular}{lllll}
    \toprule
    \textbf{Dataset} & \textbf{Words} & \textbf{Writers} & \textbf{Period} & \textbf{Language}\\
    \midrule
         IAM~\cite{marti2002iam}    & 115,320 & 657 & Modern & English\\
         GW~\cite{lavrenko2004holistic}     & 4,860 & 1 & Historical & English\\
         Rimes~\cite{augustin2006rimes}  & 66,978 & 1,300 & Modern  & French\\
         CarCrash & 24,492 & 640 & Modern & German\\
         \bottomrule
    \end{tabular}
\end{table}

\subsection{Implementation Details}

All these experiments were run using PyTorch~\cite{paszke2017automatic} on a cluster of NVIDIA GPUs. The training was done using the Adam optimizer with an initial learning rate of $2 \cdot 10^{-4}$ and a batch size of 32. We have set the dropout probability to be 50\% for all the GRU layers except the last layer of both the encoder and the decoder. There is a probability of 50\% to apply data augmentation on the training set, and we use label smoothing~\cite{szegedy2016rethinking} as a regularization mechanism.

\subsection{Ablation Study}
\label{sec:ablation}

The first experiment corresponds to an ablation study, which has been performed using the IAM dataset. The CER (\%) and WER (\%) shown correspond to the validation set of the IAM. The only one exception is Table~\ref{tab:lm}, which is applied on the GW dataset.

Firstly, different popular CNN models have been evaluated in Table~\ref{tab:cnn}. Given that the VGG19-BN model obtains the best results, we have chosen it as the feature extractor in our architecture.

\begin{table}
    \centering
    \caption{Comparison among the popular CNN models.}
    \label{tab:cnn}
    \begin{tabular}{lcc}
    \toprule
    \textbf{Model} & \textbf{CER} & \textbf{WER}\\
    \midrule
    VGG11-BN & 7.35 & 20.91\\
    VGG13-BN & 6.85 & 19.76\\
    VGG16-BN & 6.57 & 17.04\\
    \textbf{VGG19-BN} & \textbf{5.05} & \textbf{13.84}\\
    ResNet18 & 6.72 & 16.13\\
    ResNet34 & 5.51 & 14.25\\
    ResNet50 & 5.27 & 13.95\\
    ResNet101 & 5.38 & 14.34\\
    ResNet152 & 5.13 & 13.89\\
    SqueezeNet 1.0 & 6.82 & 17.35\\
    SqueezeNet 1.1 & 8.25 & 20.56\\
    Densenet121 & 5.29 & 13.79\\
    Densenet169 & 5.30 & 14.23\\
    Densenet201 & 5.37 & 13.79\\
    \bottomrule
    \end{tabular}
\end{table}{}

Secondly, we compare two different architectures of the encoder, as explained in Section~\ref{sec:enc}. The CNN+BGRU architecture obtains better results than when using a CNN with positional encoding, as shown in the Table~\ref{tab:enc}, because a trainable BGRU can provide not only the positional information, but also better mutual information among all the feature vector $\mathcal{H}$.

\begin{table}
    \centering
    \caption{Comparison between the conventional decoder unit and the proposed simplified decoder unit.}
    \label{tab:enc}
    \begin{tabular}{lcc}
    \toprule
    \textbf{Encoder} & \textbf{CER} & \textbf{WER}\\
    \midrule
    Pos. enc. & 5.67 & 14.79\\
    \textbf{CNN+BGRU} & \textbf{5.05} & \textbf{13.84}\\
    \bottomrule
    \end{tabular}
\end{table}{}

Thirdly, we compare our proposed decoder unit with the conventional decoder unit explained in Section~\ref{sec:dec}. From Table~\ref{tab:dec} we notice that the proposed decoder unit has a similar and even slightly better performance even without the post feeding of the context vector. Thus, we opt to keep the simpler version of the decoder architecture.

\begin{table}
    \centering
    \caption{Comparison between the conventional decoder unit and the proposed simplified decoder unit.}
    \label{tab:dec}
    \begin{tabular}{lcc}
    \toprule
    \textbf{Decoder Unit} & \textbf{CER} & \textbf{WER}\\
    \midrule
    Conventional & 5.06 & 13.91\\
    \textbf{Proposed} & \textbf{5.05} & \textbf{13.84}\\
    \bottomrule
    \end{tabular}
\end{table}{}

Table~\ref{tab:attn} shows the comparison between the two attention methods, as detailed in Section~\ref{sec:attn}. We can observe that location-based attention with label smoothing obtains the best performance. 

\begin{table}
    \centering
    \caption{Comparison between content and location-based attention mechanisms.}
    \label{tab:attn}
    \begin{tabular}{lccc}
        \toprule
        \textbf{Attention} & \textbf{LabelSmooth} & \textbf{CER} & \textbf{WER}\\
        \midrule
        \multirow{2}{*}{\centering Content} & $-$ & 5.79 & 15.91\\
         & \checkmark & 5.08 & 13.88\\
        \midrule
        \multirow{2}{*}{\centering Location} & $-$ & 5.49 & 14.74\\
         & \checkmark & \textbf{5.01} & \textbf{13.61}\\
         \bottomrule
    \end{tabular}
\end{table}{}

Finally, to make the best of a language model, we have investigated which is the best way to inject the linguistic knowledge. In Table~\ref{tab:act}, we have applied different activation functions on the output of the language model using the GW dataset. The best performance has been achieved without the usage of activation function while doing batch normalization on the concatenation of the three components: the prediction of external language model, the embedding of the character that is predicted by the decoder at previous time step, and the current context vector. The softmax approach is not working well because it gives too strong hypothesis to only one specific character in the available list. On the contrary, the sigmoid approach gives independent probabilities across the available character list, but it also highlights the unrelated characters. The embedding approach selects the best hypothesis from the language model and feeds its embedded format into the decoder. This can help because the embedding process has projected the relevant linguistic characters into a common latent space, which gives the decoder an opportunity to select a possible character in a closed range in that space, but the embedding process loses some useful information. Thus, the best way is to use what it is provided from the language model without any activation function, while batch normalizing the three inputs of the decoder can further improve the performance because of the similar value range for the three different vectors. 

\begin{table}
    \centering
    \caption{Comparison of the activation functions to inject the external language model.}
    \label{tab:act}
    \begin{tabular}{lcc}
        \toprule
        \textbf{Activation Function} & \textbf{CER} & \textbf{WER}\\
        \midrule
        Baseline & 2.82 & 7.13\\
        \midrule
        Softmax & 2.78 & 7.13\\
        Sigmoid & 2.81 & 7.04\\
        Embedding & 2.78 & 7.13\\
        No activation & 2.58 & 6.78\\
        \textbf{No activation + batch norm.} & \textbf{2.52} & \textbf{6.61}\\
        \bottomrule
    \end{tabular}
\end{table}{}
\subsection{Main Results}
In this section, firstly, we describe the comprehensive experiments that have been conducted to explore the best sequence-to-sequence architecture for handwritten word recognition tasks. Secondly, based on the baseline model that has been selected, we carry on further experiments with the sequence-to-sequence model equipped with the different language models to prove their different effectiveness and robustness. Finally, a real industrial use case is shown to demonstrate its applicability to industry. 

\subsubsection{Baseline Model}
\begin{figure}
\centering
\begin{tabular}{ccccc}
    \includegraphics[height=6cm]{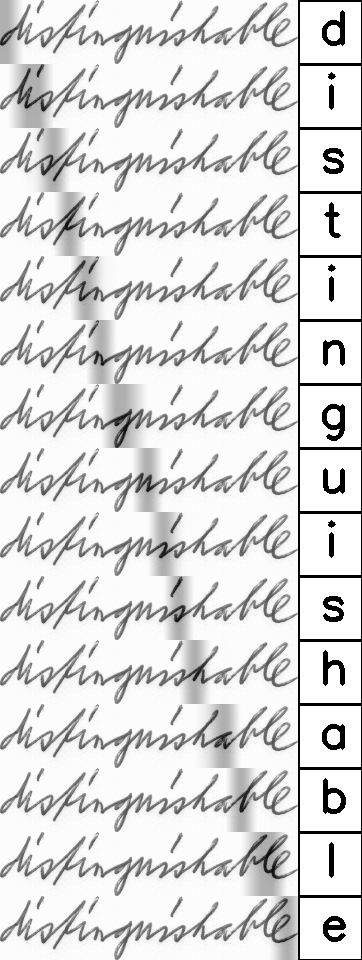} &
    \includegraphics[height=6cm]{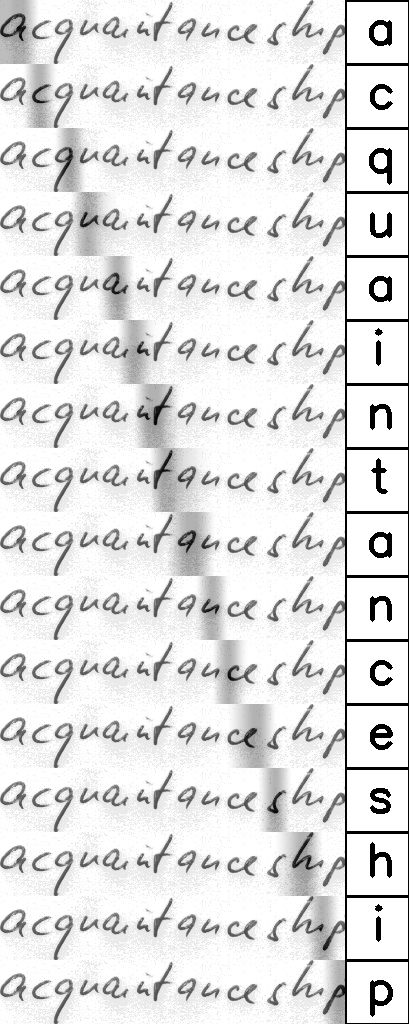} &
    \includegraphics[height=6cm]{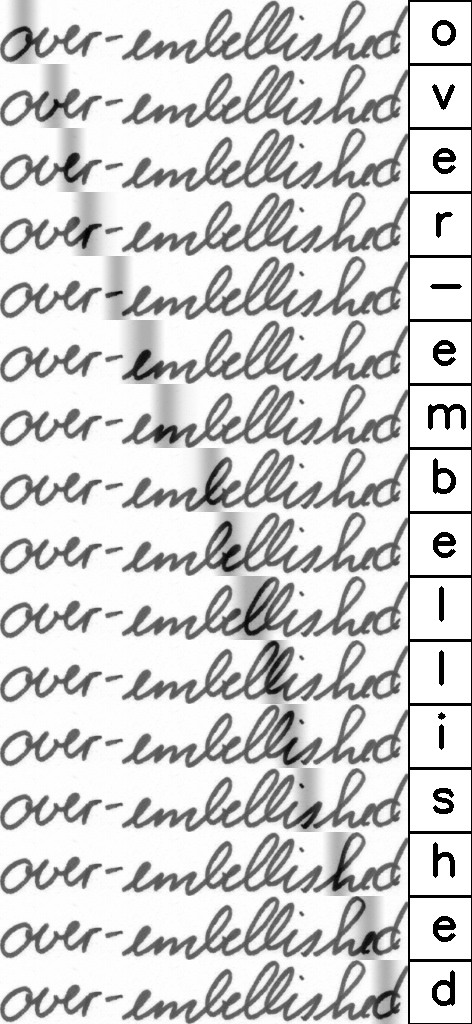} &
    \includegraphics[height=6cm]{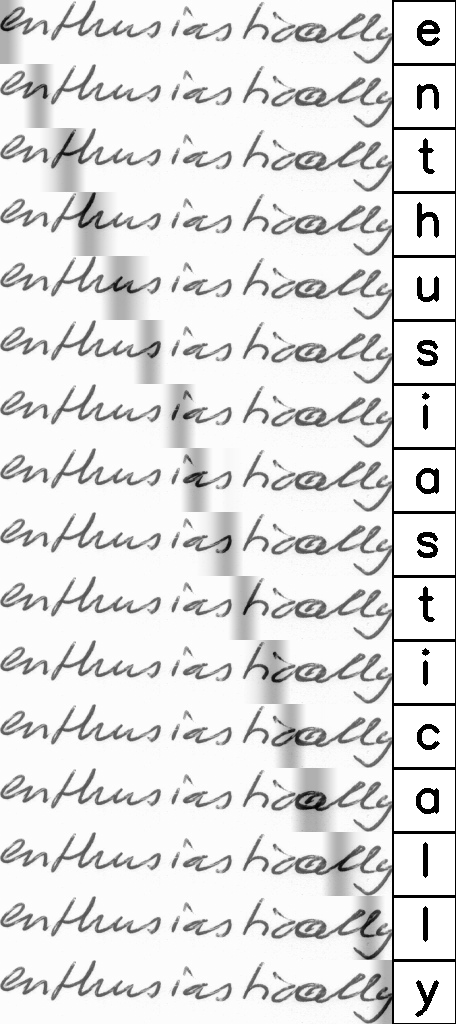} &
    \includegraphics[height=6cm]{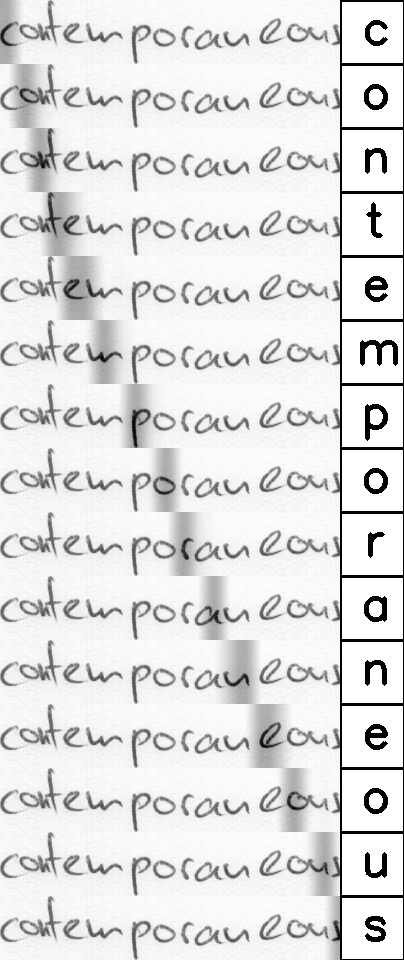}\\
\end{tabular}
\caption{Visualization samples of attention on the IAM dataset.}
\label{fig:attn1}
\end{figure}

\begin{figure}
\centering
\begin{tabular}{ccccc}
    \includegraphics[height=5cm]{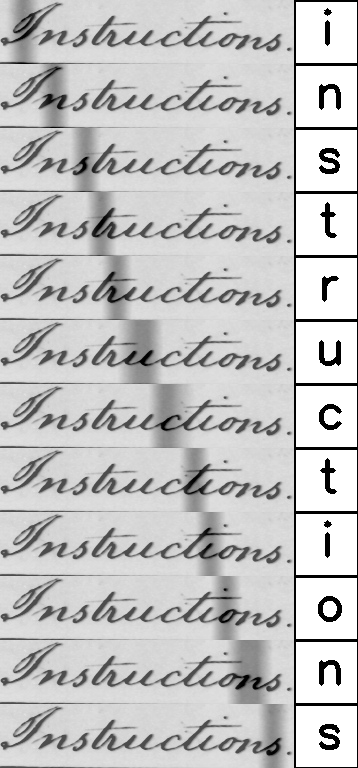} &
    \includegraphics[height=5cm]{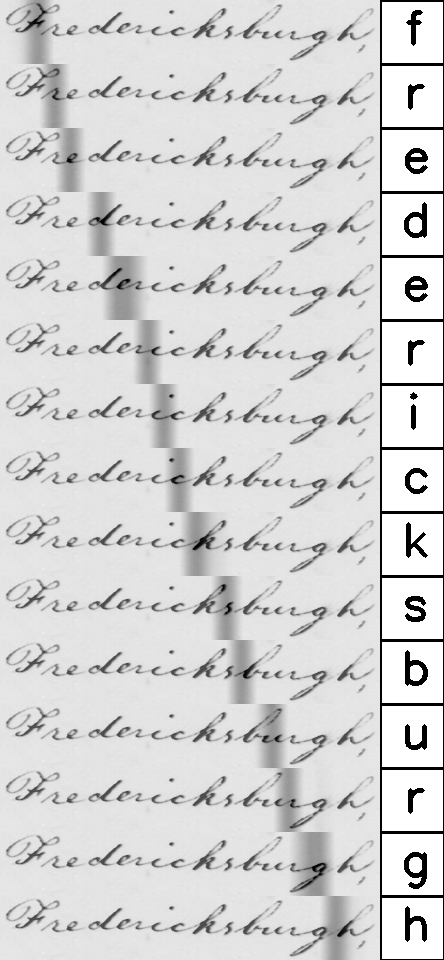} &
    \includegraphics[height=5cm]{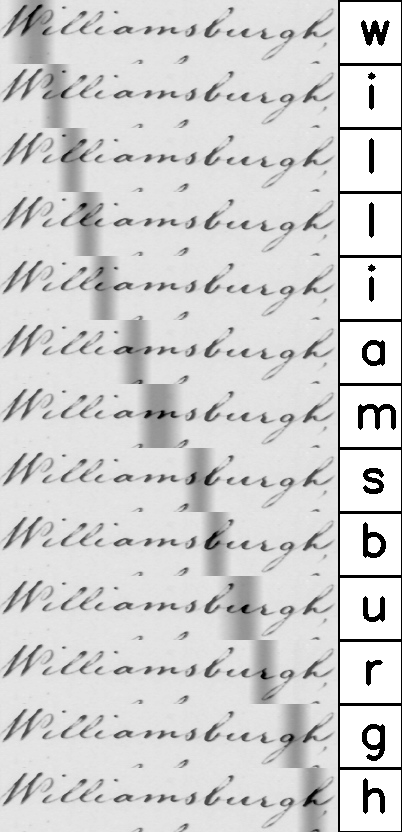} &
    \includegraphics[height=5cm]{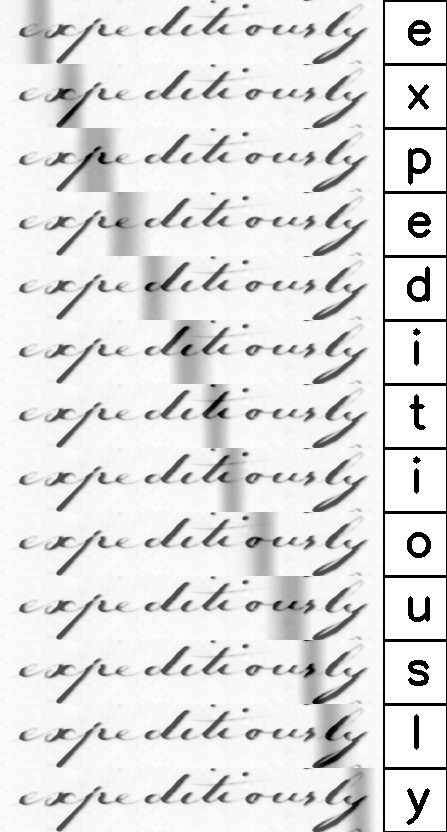} &
    \includegraphics[height=5cm]{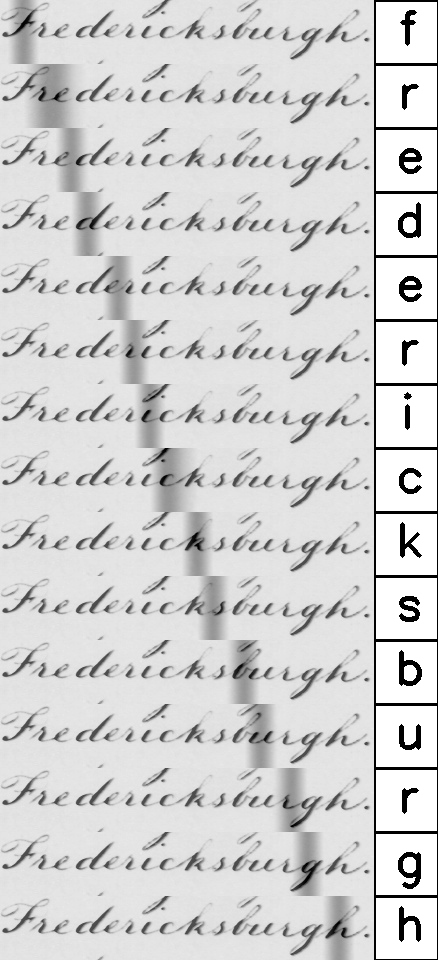}
\end{tabular}
\caption{Visualization samples of attention on the GW dataset.}
\label{fig:attn2}
\end{figure}

\begin{figure}
\centering
\begin{tabular}{cccc}
    \includegraphics[height=6cm]{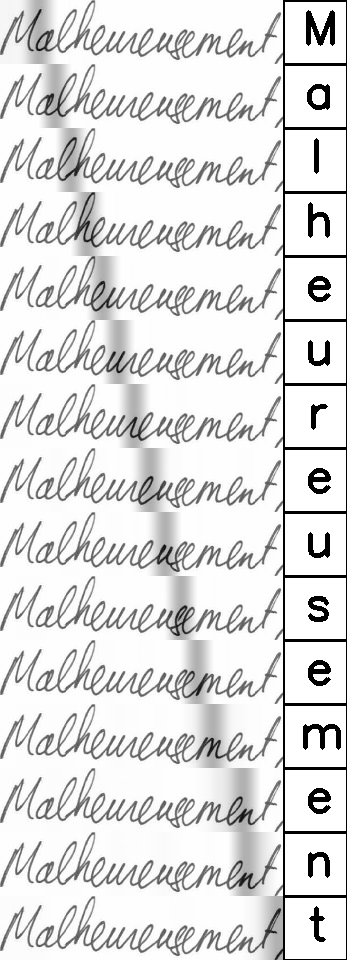} &
    \includegraphics[height=6cm]{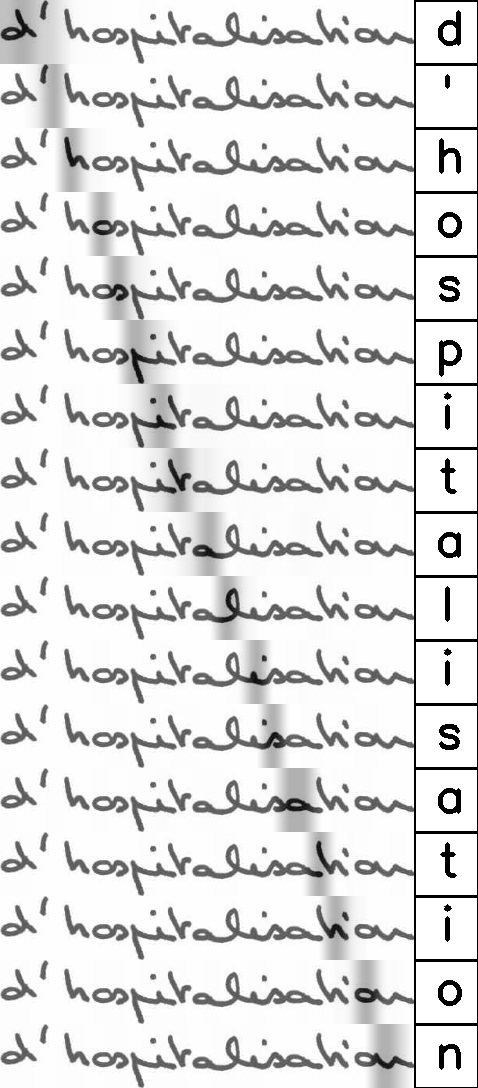} &
    \includegraphics[height=6cm]{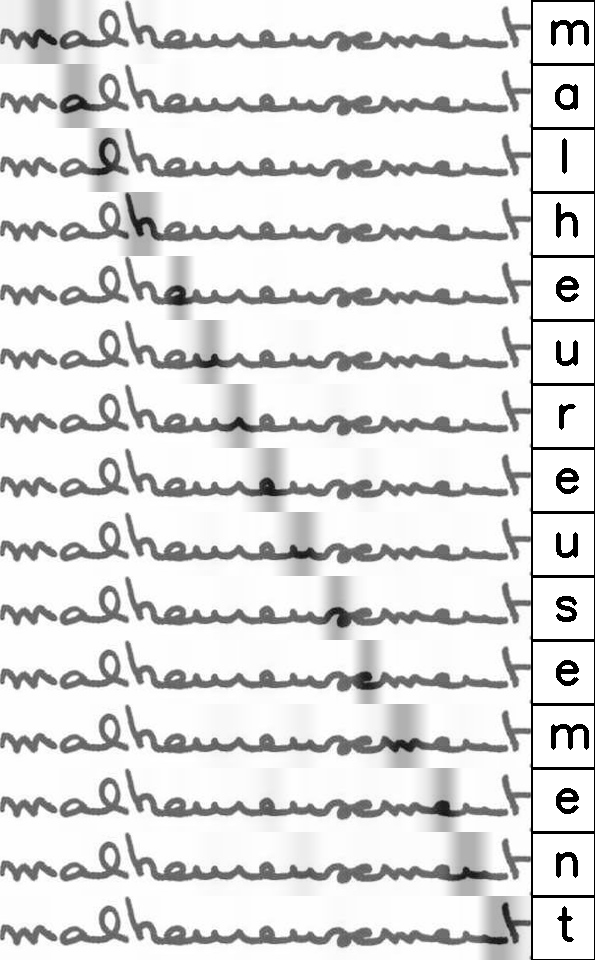} &
    \includegraphics[height=6cm]{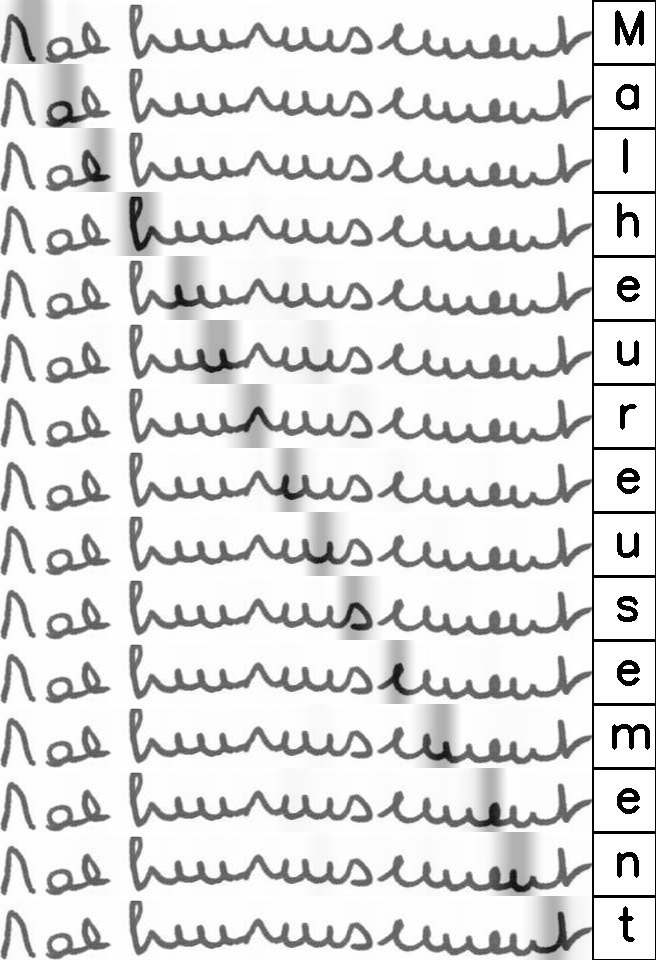}
\end{tabular}
\caption{Visualization samples of attention on the Rimes dataset.}
\label{fig:attn3}
\end{figure}

We would like to analyze the performance of our sequence-to-sequence recognizer without any assistance from external language model nor a lexicon. So, in Table~\ref{tab:baseline}, we have listed all the comparable results achieved by the state-of-the-art handwriting recognizers. From the table, we observe that our recognizer achieves good performance on the IAM, GW and Rimes datasets. We show some examples of the visualized attention maps on the IAM (Figure~\ref{fig:attn1}), GW (Figure~\ref{fig:attn2}) and Rimes (Figure~\ref{fig:attn3}). From those examples, we observe that the attention is able to attend each character at its corresponding time step. In addition, it can adapt itself to change its focus depending on the varied width of each character. 

\begin{table*}
\centering
\caption{Comparison with the state-of-the-art handwritten word recognition works, without language model nor lexicon.}
\label{tab:baseline}
\begin{minipage}{\linewidth}
\centering
\begin{tabular}{lcccccccc}
\toprule
 & \multicolumn{2}{c}{IAM} &&\multicolumn{2}{c}{GW} && \multicolumn{2}{c}{Rimes}\\
\cmidrule{2-3}\cmidrule{5-6}\cmidrule{8-9}
Method & CER & WER && CER & WER && CER & WER\\
\midrule
Mor~\etal~\cite{mor2018confidence} & $-$ &20.49 && $-$ &  $-$ &&  $-$ & 11.95\\
Pham~\emph{et al.}~\cite{pham2014dropout} & 13.92 & 31.48&& $-$ & $-$ && 8.62 & 27.01 \\
Bluche~\emph{et al.}~\cite{bluche2017scan} & 12.60 & $-$ && $-$ & $-$ && $-$ & $-$ \\
Wiginton~\emph{et al.}~\cite{wigington2017data} & 6.07 & 19.07&& $-$ & $-$ && 3.09 & 11.29 \\
Sueiras~\emph{et al.}~\cite{sueiras2018offline} & 8.80 & 23.80 && $-$ & $-$ && 4.80 & 15.90  \\
Kang~\emph{et al.}~\cite{kang2018convolve} & 6.88 & 17.45 && $-$ & $-$ && $-$ & $-$ \\
Krishnan~\emph{et al.}~\cite{krishnan2018word} & 6.34 & 16.19 && $-$ & $-$ && $-$ & $-$\\
Toledo~\emph{et al.}~\cite{toledo2017handwriting} & $-$ & $-$ && 7.32 & $-$ && $-$ & $-$ \\
Dutta~\etal~\cite{dutta2018improving}\footnote{This work provides the results using \textbf{Test-time Augmentation}, which are not directly comparable with other results.} & \textbf{4.88} & \textbf{12.61} && 4.29 & 12.98 && \textbf{2.32} & \textbf{7.04}\\
\midrule
\textbf{Proposed} & 5.79 & 15.15 && \textbf{2.82} & \textbf{7.13} && 2.65 & 8.71\\
\bottomrule
\end{tabular}
\end{minipage}
\end{table*}

\subsubsection{Integration of the Language Model}
In this subsection we evaluate the performance of our language model by expanding the baseline model shown in Table~\ref{tab:baseline}. The results are shown in Table~\ref{tab:lm}. Our language model is pre-trained with a large corpus, detailed in Section~\ref{sec:syn}, and fine-tuned with the training data within the whole sequence-to-sequence system during end-to-end training process. Thus, the language model can adapt itself to a specific dataset corpus while keeping the capacity of generalization.

As we can see in this table, our language model can boost the performance on all the three datasets, achieving better results than the Shallow Fusion and Deep Fusion language models. In fact, the Shallow Fusion makes the performance to decrease on all the three datasets, because it is too sensitive that any peaky probability distribution from both the outputs of the decoder and the external language model can ruin the final result. The Deep Fusion model behaves quite well on the GW and Rimes datasets, being able to improve the results a little bit compared to the baseline. In conclusion, our proposed Candidate Fusion is better than the Shallow and Deep Fusion approaches, because it is trainable and flexible to assist the recognizer during decoding. In addition, it does not need to manually tune the trade-off between the outputs of decoder and language model. In fact, in our Candidate Fusion architecture, the role of an external language model is to provide an extra predicted transcription based on the recognizer's prediction and its own language knowledge, while at the same time, the external language model can be adapted to the most commonly errors made by the sequence-to-sequence optical recognizer.


\begin{table*}
\centering
\caption{Comparison with the state-of-the-art handwritten word recognition with language model, but not constrained by a lexicon.}
\label{tab:lm}
\resizebox{\columnwidth}{!}{
\begin{tabular}{lcccccccc}
\toprule
 & \multicolumn{2}{c}{IAM} &&\multicolumn{2}{c}{GW} && \multicolumn{2}{c}{Rimes}\\
\cmidrule{2-3}\cmidrule{5-6}\cmidrule{8-9}
Method & CER & WER && CER & WER && CER & WER\\
\midrule
Baseline no LM & 5.79 & 15.15 && 2.82 & 7.13 && 2.65 & 8.71\\
\midrule
Shallow Fusion LM & 6.14 & 16.12 && 2.95 & 7.73 && 3.63 & 12.29\\
Deep Fusion LM & 5.91 & 15.45 && 2.72 & 6.79 && 2.54 & 8.20\\
\midrule
\textbf{Proposed Candidate Fusion LM} & \textbf{5.74} & \textbf{15.11} && \textbf{2.52} & \textbf{6.61} && \textbf{2.32} & \textbf{7.47}\\
\bottomrule
\end{tabular}
}
\end{table*}

\subsubsection{Restriction with a Close Dictionary}
In all the experiments shown above, we never restrict the recognizer to a specific lexicon, which means the recognizer can predict out-of-vocabulary (OOV) words. Indeed, a generic handwritten word recognizer should not be restricted to closed lexicon in industrial use cases. However, since the use of closed lexicons is also a common practice, we have also tested how it can improve the overall performance. Thus, in Table~\ref{tab:lexicon}, we have applied a simple edit-distance method to find the closest word in three lexicons: the brown lexicon with the lexicon of the test set (te+brown), the lexicon from the target dataset (tr+va+te), and only the lexicon of the test set (te). As expected, a lexicon can always improve the performance. 

\begin{table}
\centering
\caption{Results after applying a simple edit-distance based lexicon constraint.}
\label{tab:lexicon}
\begin{tabular}{lcccccccc}
\toprule
 & \multicolumn{2}{c}{IAM} &&\multicolumn{2}{c}{GW} && \multicolumn{2}{c}{Rimes}\\
\cmidrule{2-3}\cmidrule{5-6}\cmidrule{8-9}
Lexicon & CER & WER && CER & WER && CER & WER\\
\midrule
Baseline & 5.74 & 15.11 && 2.52 & 6.61 && 2.40 & 7.74\\
\midrule
te+brown & 5.08 & 10.51 && 2.29 & 5.07 && 1.88 & 4.53\\
tr+va+te &	4.51 & 8.99 && 1.84 & 3.86 && 1.75 & 4.45\\	
\textbf{te} & \textbf{4.27} & \textbf{8.36} && \textbf{1.88} & \textbf{3.61} && \textbf{1.45} & \textbf{3.75}\\	
\bottomrule
\end{tabular}
\end{table}

\subsubsection{Application to a Real Industrial Use Case}
\begin{figure*}
    \centering
    \includegraphics[width=0.99\linewidth]{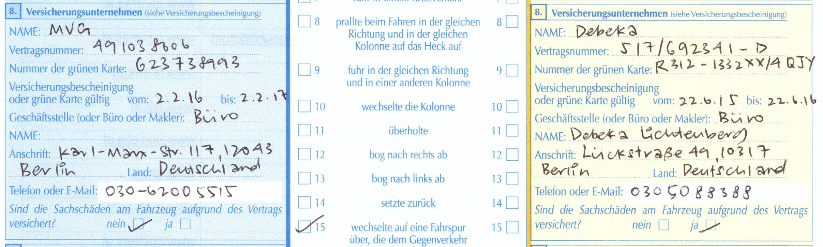}
    \caption{A cropped area of the real industrial use case dataset.}
    \label{fig:real}
\end{figure*}

Finally, we evaluate our recognizer in a real world scenario for recognizing handwritten fields in car crash statement forms. For this experiment, we have applied our system to an in-house private dataset. Due to the privacy protection, we could only show a cropped image of the real dataset in Figure~\ref{fig:real}. This real dataset consists of numbers, words, sentences and even check-boxes, which are way more challenging than the popular scientific datasets shown above. Compared with a well-known CTC-based approach~\cite{puigcerver2017multidimensional}, our proposed approach achieves better performance, as shown in Table~\ref{tab:real}. This results suggests that our model has a good generalization ability. 
 
\begin{table}
	\centering
	\caption{Results of a real use case.}
	\label{tab:real}
	\begin{tabular}{lcc}
		\toprule
		Method & CER & WER\\
		\midrule
		CTC-based~\cite{puigcerver2017multidimensional} & 5.6 & 7.4\\
		\textbf{Proposed} & \textbf{3.7} & \textbf{4.5}\\
		\bottomrule
	\end{tabular}
\end{table}

\section{Conclusion}
\label{sec:con}
In this paper we have proposed a novel way to integrate an external language model into a sequence-to-sequence model for handwritten word recognition. Our language model approach, named Candidate Fusion, is trained and optimized together with the optical recognizer, so that we can avoid biases between different training corpuses, and has the advantage that the language model guides the decoding according to the most likely character sequence. The extensive evaluation, including an ablation study as well as comparisons with state-of-the-art approaches, demonstrates the effectiveness of our approach. Indeed it not only outperforms the existing approaches on public scientific datasets, but it also shows its robustness on a real industrial use case.


\section*{Acknowledgments}

This work has been partially supported by the Spanish project RTI2018-095645-B-C21, the grant 2016-DI-087 from the Secretaria d'Universitats i Recerca del Departament d'Economia i Coneixement de la Generalitat de Catalunya, the grant FPU15/06264 from the Spanish Ministerio de Educaci\'{o}n, Cultura y Deporte, the Ramon y Cajal Fellowship RYC-2014-16831 and the CERCA Program/ Generalitat de Catalunya. We gratefully acknowledge the support of NVIDIA Corporation with the donation of the Titan Xp GPU used for this research.

\bibliographystyle{elsarticle-num.bst}

\bibliography{bib.bib}





\end{document}